# Detection of copy-move forgery in digital images based on DCT

Nathalie Diane Wandji[1], Sun Xingming[2], Moise Fah Kue[3]

[1] School of Information Science and Engineering, Hunan University
Changsha, Hunan, P.R China

[2] Jiangsu Engineering Center of Network Monitoring, Nanjing University of Information Science and Technology
Nanjing, Jiangsu, P.R China

[3] Independent scholar
Bertoua, East, Cameroon

**Abstract**

With rapid advances in digital information processing systems, and more specifically in digital image processing software, there is a widespread development of advanced tools and techniques for digital image forgery. One of the techniques most commonly used is the Copy-move forgery which proceeds by copying a part of an image and pasting it into the same image, in order to maliciously hide an object or a region. In this paper, we propose a method to detect this specific kind of counterfeit. Firstly, the color image is converted from RGB color space to YCbCr color space and then the R, G, B and Y-component are splitted into fixed-size overlapping blocks and, features are extracted from the R, G and B-components image blocks on one hand and on the other, from the DCT representation of the R, G, B and Y-component image block. The feature vectors obtained are then lexicographically sorted to make similar image blocks neighbors and duplicated image blocks are identified using Euclidean distance as similarity criterion. Experimental results showed that the proposed method can detect the duplicated regions when there is more than one copy move forged area in the image and even in case of slight rotations, JPEG compression, shift, scale, blur and noise addition.

## 1. Introduction

Nowadays, with the wide availability of powerful and easy-to-use image processing softwares, even a non-professional can tamper with an image without leaving any trace detectable by the human vision system. Consequently, the integrity of digital images can no longer be presumed without further investigations. In order to regain image's credibility especially when it comes to sensitive data such as evidences in court of law, news items, medical records to name but few, reliable techniques must be developed to examine the integrity and/or authenticity of digital images.

Generally, these techniques are divided into two major categories: the active methods on one hand and the passive/blind methods on the other hand. The drawback of the first category is the requirement that certain information is embedded into the image either during its creation or before its broadcasting to the public. Digital watermarking belongs to this category [1]. The inserted information can be used either to detect the source of an image or to detect possible modifications of an image. On the contrary, passive / blind methods do not require any prior information to be embedded into the digital image. They work with the assumption that any forgery, even if not visible with naked eyes, would modify the intrinsic statistics of the original image. Copy-move forgery belongs to this category [2]. As mentioned previously, this kind of forgery is mostly used to hide a specific object or area in the original image.

Several methods have been developed to detect copy-move forgeries. In [3], Fridrich et al first described the exhaustive search indicating that its applicability is limited mainly because of its exponential complexity and the fact that it fails in case of any distorsion. In the same paper, they proposed a more effective approach, which uses a robust representation of the block that consists of quantized discrete cosine transform (DCT) coefficients. Popescu et al. proposed a resembling method [4], which used principal component analysis (PCA) instead of DCT to generate the block representation. They went further by reducing by half the numbers of features used in [3] and therefore improving the efficiency. Despite these improvements, their method has some weaknesses, among which its failure in case of slight rotation of the copied region. Later, Weiqi Luo et al. [5] presented a technique robust to various forms of post region duplication processing, including blurring, noise contamination and lossy compression. They represented each block by 7 characteristics extracted from both the RGB color image and the YCbCr corresponding image. Li Kang et al. [6] suggested to apply improved singular value decomposition to each image block to yield a reduced dimension representation and then lexicographically sort the feature matrix formed by the singular values. Their method was proven to be robust against noise distortion. Weihai Li et al. [7] proposed a rotation-robust algorithm based on the Fourier-Mellin Transform of image's blocks with features extracted along radius direction. Recently, Y. Huang et al. [8] proposed an improved DCT-based method. In their approach, DCT is applied to each block to represent its features and then truncating it yields a reduced dimension representation of the features. Their method has been proven to be robust to JPEG compression, blurring or AWGN distortions but they failed to consider the multiple copy-move forgery. Most recently, Yanjun Cao et al. [9] proposed an approach based on improved DCT that has the advantages to be robust to various attacks, such as multiple copy-move forgery, Gaussian blurring, and noise contamination; and also to have a lower computational complexity.

In this paper, we propose a DCT-based approach, which is not only robust to multiple copy-move forgery, Gaussian blurring, and noise contamination, but also to rotation with an angle up to 5 degrees. The rest of the paper is organized as follows: Section 2 will describe the proposed approach and Section 3 will present the experimental results. Conclusion is drawn in Section 4.

## 2. The proposed Method

In copy-move forgery, since the copied regions come from the same image, at the end of the process, we will have relatively similar areas in the image. The detection of such forgery will therefore consist in finding wide relatively similar areas in an image. The easiest way to detect those areas is the exhaustive search but this can only be done for very small images because it is computationally costly. Moreover, it fails when the copied region is further processed. To make the detection more efficient, we will use the most common approach that starts by dividing the suspected image into overlapping blocks. Once the division is done, robust features must be extracted from the blocks in order to have an efficient detection rate. At last, the features are sorted to make a sufficiently reliable decision based on the similarity of consecutive pairs.

### 2.1 Overview of the Method

The different steps of our method are presented as follows:
1) Convert the RGB image to YUV color space.
2) Divide the R, G, B and Y-component into fixed-sized blocks.
3) Transform each block into DCT coefficients.
4) Extract features from the obtained DCT coefficients and the Red, Green and Blue channels
5) Sort the features in lexicographic order.
6) Search for similar pairs of blocks.
7) Output the duplicated regions if any.

### 2.2 Detailed Procedure

The algorithm described in 2.1 is implemented as follows.
*(i)* The suspicious RGB *color* image f of size $M \times N \times 3$ is first converted to YUV color space as described in Eq. 1.

$$\begin{cases} Y = 0.299 * R + 0.587 * G + 0.114 * B \\ U = -0.14713 * R - 0.28886 * G + 0.436 * B \\ V = 0.615 * R - 0.51498 * G - 0.10001 * B \end{cases} \quad (1)$$

*(ii)* Then the R, G, B and Y-component $f_R$, $f_G$, $f_B$, $f_Y$ are splitted each into overlapping square fixed-size $b \times b$ blocks (here $b=16$), generating $(M-b+1)(N-b+1)$ blocks per component and therefore a total of $(M-b+1)(N-b+1) \times 4$ blocks. The obtained blocks are denoted as $R_{ij}$, $G_{ij}$, $B_{ij}$ and $Y_{ij}$ respectively where $i$ and $j$ indicates the starting point of the block's row and column, respectively.

$$\begin{cases} R_{ij}(x,y) = f_R(x+i, y+j), \\ G_{ij}(x,y) = f_G(x+i, y+j), \\ B_{ij}(x,y) = f_B(x+i, y+j), \\ Y_{ij}(x,y) = f_Y(x+i, y+j), \end{cases} \quad (2)$$

$x, y \in \{0, ..., b-1\}$, $i \in \{0, ..., M-b\}$, and $j \in \{0, ..., N-b\}$

*(iii)* DCT coefficients are computed over each $b \times b$ block as follows for :

$$\begin{cases} CR_{ij}(x,y) = DCT(f_R(x+i, y+j)), \\ CG_{ij}(x,y) = DCT(f_G(x+i, y+j)), \\ CB_{ij}(x,y) = DCT(f_B(x+i, y+j)), \\ CY_{ij}(x,y) = DCT(f_Y(x+i, y+j)), \end{cases} \quad (3)$$

$x, y \in \{0, ..., b-1\}$, $i \in \{0, ..., M-b\}$, and $j \in \{0, ..., N-b\}$

*(iv)* Features are extracted as follows.
$V_{ij} = [V_{ij}^1, V_{ij}^2, V_{ij}^3, V_{ij}^4, V_{ij}^5, V_{ij}^6, V_{ij}^7, V_{ij}^8, V_{ij}^9]$
where the elements are defined as described in Eq. 4.

$$\begin{cases} V_{ij}^1 = CY_{ij}(0,0) \\ V_{ij}^2 = CY_{ij}(0,1) \\ V_{ij}^3 = CY_{ij}(1,0) \\ V_{ij}^4 = CR_{ij}(0,0) \\ V_{ij}^5 = CG_{ij}(0,0) \\ V_{ij}^6 = CB_{ij}(0,0) \\ V_{ij}^7 = \text{Average}_{0 \leq x \leq M-b+1, 0 \leq y \leq N-b+1}(R_{ij}(x,y)) \\ V_{ij}^8 = \text{Average}_{0 \leq x \leq M-b+1, 0 \leq y \leq N-b+1}(B_{ij}(x,y)) \\ V_{ij}^9 = \text{Average}_{0 \leq x \leq M-b+1, 0 \leq y \leq N-b+1}(G_{ij}(x,y)) \end{cases} \quad (4)$$

*(v)* The extracted feature vectors are arranged into an $(M - b + 1)(N - b + 1) \times 9$ matrix, denoted as A.

*(vi)* Then the matrix A is lexicographically sorted and a matching procedure is applied: pairs of neighboring vectors are tested to decide on their similarity using their Euclidian distance. To determine the amount of neighbors for each vector, we set a threshold $T_n$.

If the distance between two neighboring vectors is smaller than a pre-defined threshold $T_1$, the analysed blocks are considered as a pair of candidate for the forgery. Their respective positions $(x_i, y_i), (x_j, y_j)$ and the shift vector between them $[|x_i - x_j|, |y_i - y_j|]$ are therefore stored.

For each pair of candidate, we compute the accumulative number of the corresponding shift vector. If that number is greater than a predefined threshold $T_s$ the corresponding regions are marked as suspicious but the decision can be final only if the involved regions are not neighbors. In fact, considering that neighboring blocks might with a high probability have similar feature vector, we need to evaluate the actual euclidian distance between both regions and make sure that distance is greater than a predefined threshold $T_2$. Only when this is done, can we conclude the regions are duplicated.

## 3. Experimental results

In this section, we evaluate the performance of the proposed method on a set of forged images generated using the Image Manipulation Program GIMP 2.6.12. The original image used was downloaded from [10] (Fig.1)

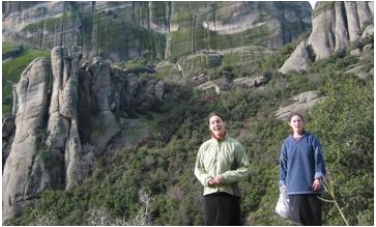

Fig. 1 Original image

We scaled it to the size $330 \times 200$ and tampered with it in 4 different manners that we will call forgery 1, forgery 2, forgery 3 and forgery 4 respectively in the rest of this paper (Fig. 2 and Fig. 3).

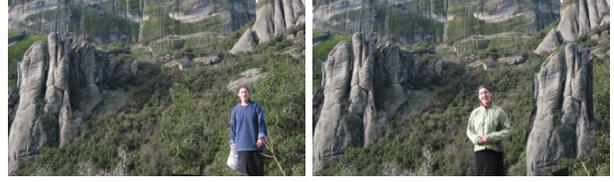

Fig. 2 Forgery 1 (Left) and Forgery 2 (Right)

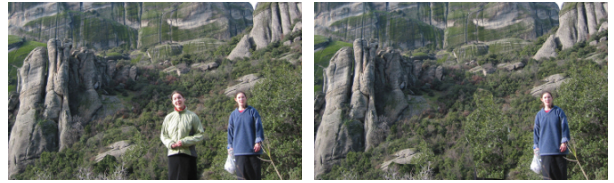

Fig. 3 Forgery 3 (Left) and Forgery 4 (Right)

We have evaluated the robustness of our algorithm against JPEG compression, noise addition, scaling, rotation, blurring and shift. The experimental results presented were all done with the values $b=16$ (block size), $T_1=0.014$; $T_2=2*b=32$; $T_s=b+2=18$; $T_n=b=16$.

The detection rate $d$ and false detection rate $f$ are defined in Eq. 7

$$d = \frac{|C \cap D|}{|C|}, f = \frac{|D \setminus C|}{|C|} \quad (5)$$

where $C$ represents the actual copy-move tampered area; $D$ represents the detected area; and $|C|$ represents the total number of pixels in $C$. $D \setminus C = \{x \epsilon D | x \notin C\}$. Both rates will be expressed in percentage (%).

### 3.1 JPEG Compression

JPEG compression was applied forged images with quality factors (QF) ranging from 70 to 100. (Tab.1, Fig. 4)

Table 1: Results for JPEG compression

| QF | Forgery 1 | | Forgery 2 | | Forgery 3 | |
|---|---|---|---|---|---|---|
| | d | f | d | f | d | f |
| 100 | 90.8497 | 19.8210 | 97.4390 | 17.3738 | 96.0538 | 33.4305 |
| 90 | 83.5465 | 16.6098 | 96.4080 | 13.3424 | 90.7623 | 23.8565 |
| 80 | 75.1634 | 7.8005 | 92.6967 | 10.8247 | 79.2377 | 13.4978 |
| 70 | 52.4297 | 7.7295 | 90.6891 | 7.9164 | 43.7220 | 4.2152 |

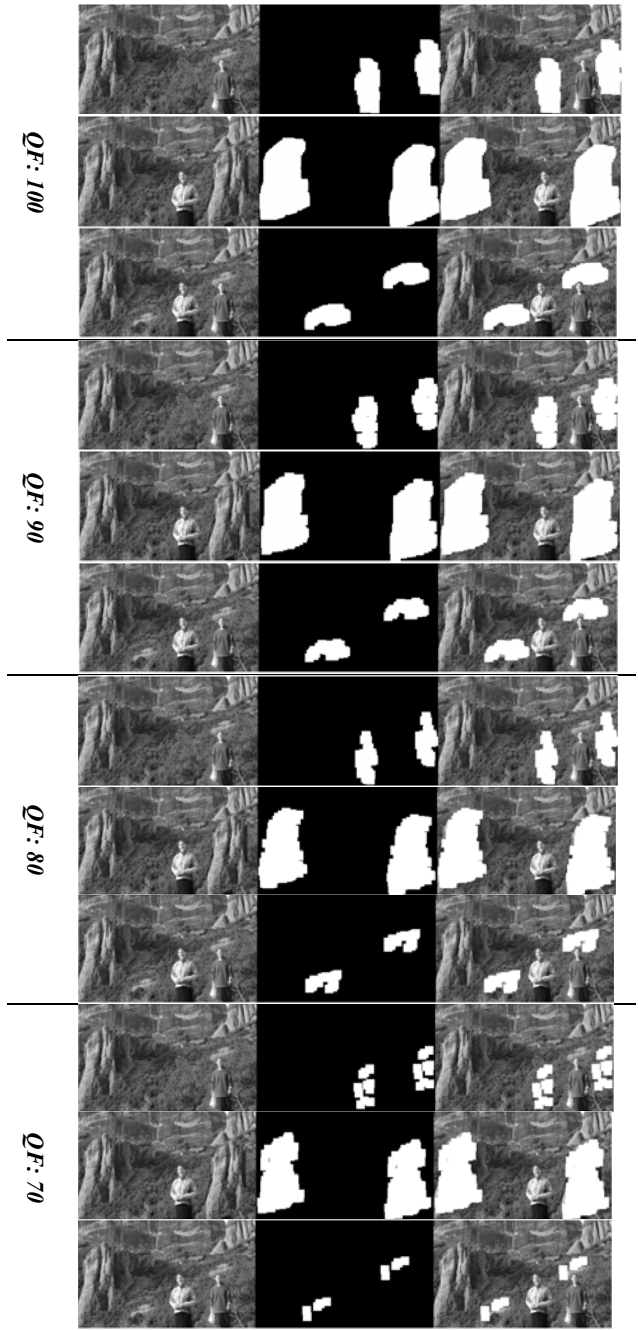

Figure 4: Results for JPEG compression

## 3.2 Gaussian Blur

We used two sets of images, one in which the copied area is blurred before pasting it; and the other in which the whole image is blurred after the forgery. Experimental results with both sets can be seen in Tab.2 and Fig. 5. They show the method is more efficient in case of blurring operation applied to the whole tampered image.

Table 2: Results for Gaussian Blurring

| | Blurring of the copied segment | | | |
|---|---|---|---|---|
| | *Forgery 2* | | *Forgery 3* | |
| radius | d | f | d | r |
| 1x1 | 89.1264 | 8.8117 | 83.2287 | 13.0717 |
| 2x2 | 73.6842 | 6.1422 | 42.1076 | 1.9731 |
| | Blurring of the whole tampered image | | | |
| | *Forgery 2* | | *Forgery 3* | |
| radius | d | f | d | f |
| 1x1 | 96.5165 | 16.4623 | 94.3946 | 35.6278 |
| 2x2 | 95.2903 | 15.5236 | 93.4081 | 28.4529 |
| 3x3 | 94.9213 | 15.0081 | 92.4215 | 45.9865 |

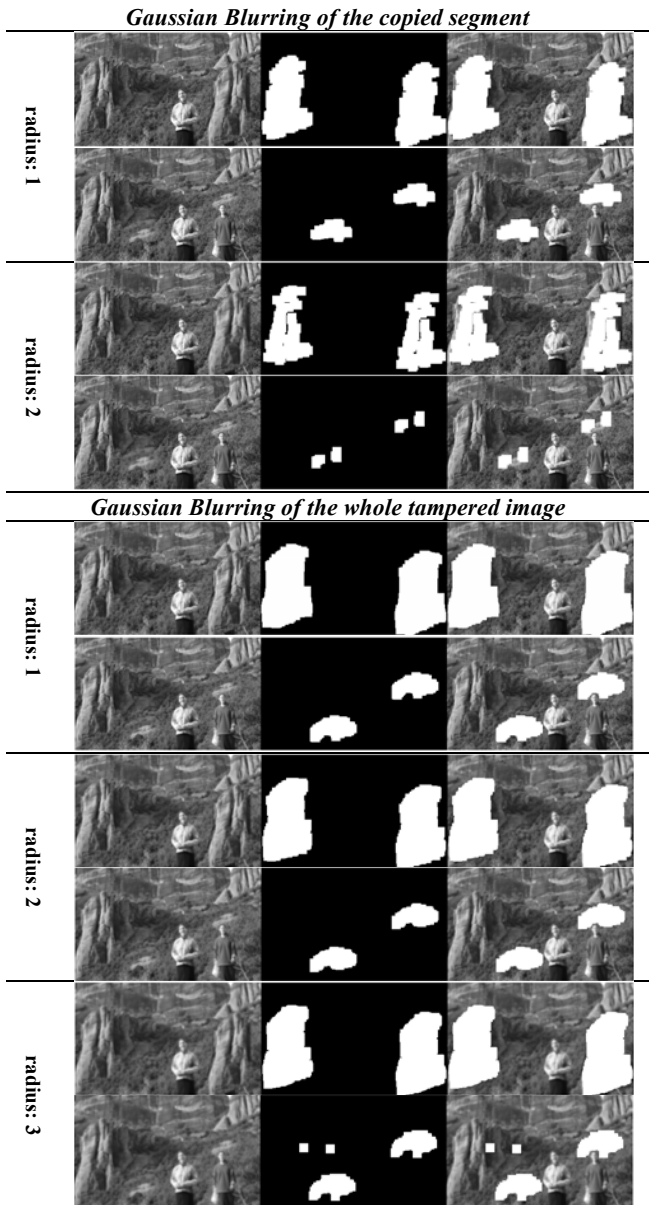

Figure 5: Results for Gaussian Blurring

Tab 3: Results for Additive Gaussian Noise

| Gaussian noise added to the copied segment | | | | |
|---|---|---|---|---|
| | Forgery 2 | | Forgery 3 | |
| *std* | *d* | *f* | *d* | *f* |
| 0.1 | 97.4390 | 17.3738 | 96.4126 | 35.4260 |
| 0.3 | 96.9940 | 17.1243 | 96.0538 | 32.6906 |
| 0.9 | 96.9181 | 16.5925 | 94.0807 | 29.6861 |
| 1.3 | 96.7553 | 16.1422 | 92.1525 | 24.2601 |
| Gaussian noise added to the whole tampered image | | | | |
| | Forgery 2 | | Forgery 3 | |
| *std* | *d* | *f* | *d* | *f* |
| 0.1 | 97.2328 | 17.1785 | 93.3632 | 27.7130 |
| 0.3 | 97.2111 | 16.5003 | 92.5561 | 26.4126 |
| 0.9 | 96.1476 | 15.9848 | 94.1704 | 30.2691 |
| 1.3 | 96.4731 | 15.8058 | 93.4081 | 23.6323 |

## 3.3 Additive Gaussian Noise

We used two sets of images as it was the case for Gaussian blurring (subsection 3.2). In both sets, noise Gaussian noise was added either on the copied segment or the full tampered image with standard deviations (*std*) between 0.1 and 1.3. In Tab. 3 and Fig. 6,

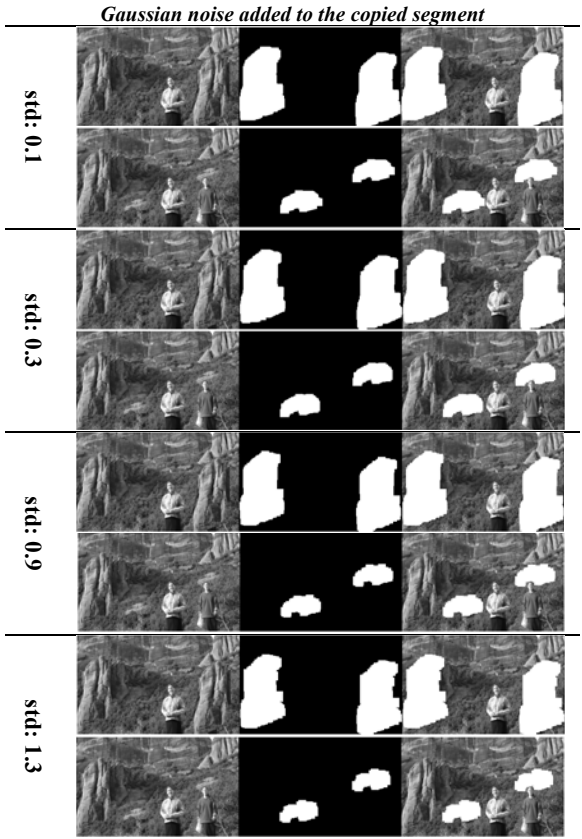

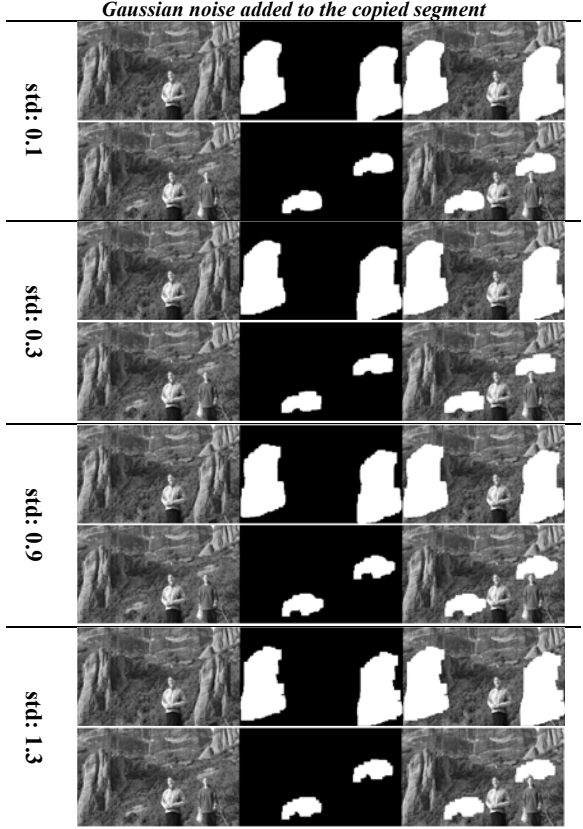

Figure 6: Results for Gaussian noise addition

## 3.4 Scaling

We used two sets of images as it was the case for Gaussian blurring (subsection 3.2). In both sets, we used a scale factor between 0.96 and 1.06. experimental results are presented in Tab. 4 and Fig. 7.

Table 4: Results for Scaling

| Scaling of the copied segment | | | | |
|---|---|---|---|---|
| | Forgery 2 | | Forgery 3 | |
| *factor* | *d* | *f* | *d* | *f* |
| 0.96 | 78.0684 | 0.8573 | 28.3857 | 3.4529 |
| 0.97 | 73.9989 | 0.3093 | 11.8033 | 0 |
| 0.98 | 65.5562 | 6.4460 | 45.2466 | 8.7720 |
| 1.02 | 74.4655 | 6.0933 | 71.7040 | 7.9821 |
| 1.04 | 64.6989 | 2.0022 | 39.8206 | 3.7220 |
| 1.06 | 53.6517 | 1.2588 | 18.9686 | 3.9910 |
| Scaling of the whole tampered image | | | | |
| | Forgery 2 | | Forgery 3 | |
| *factor* | *d* | *f* | *d* | *f* |
| 0.96 | 64.6120 | 7.0808 | 53.9462 | 9.4395 |
| 0.97 | 82.7564 | 12.1161 | 48.6996 | 6.7265 |
| 0.98 | 90.1791 | 14.8020 | 65.5605 | 25.9193 |
| 1.02 | 85.7298 | 13.6462 | 64.6188 | 19.7758 |
| 1.04 | 92.5339 | 12.1867 | 64.6188 | 18.0717 |
| 1.06 | 100 | 16.6576 | 100 | 30.6278 |

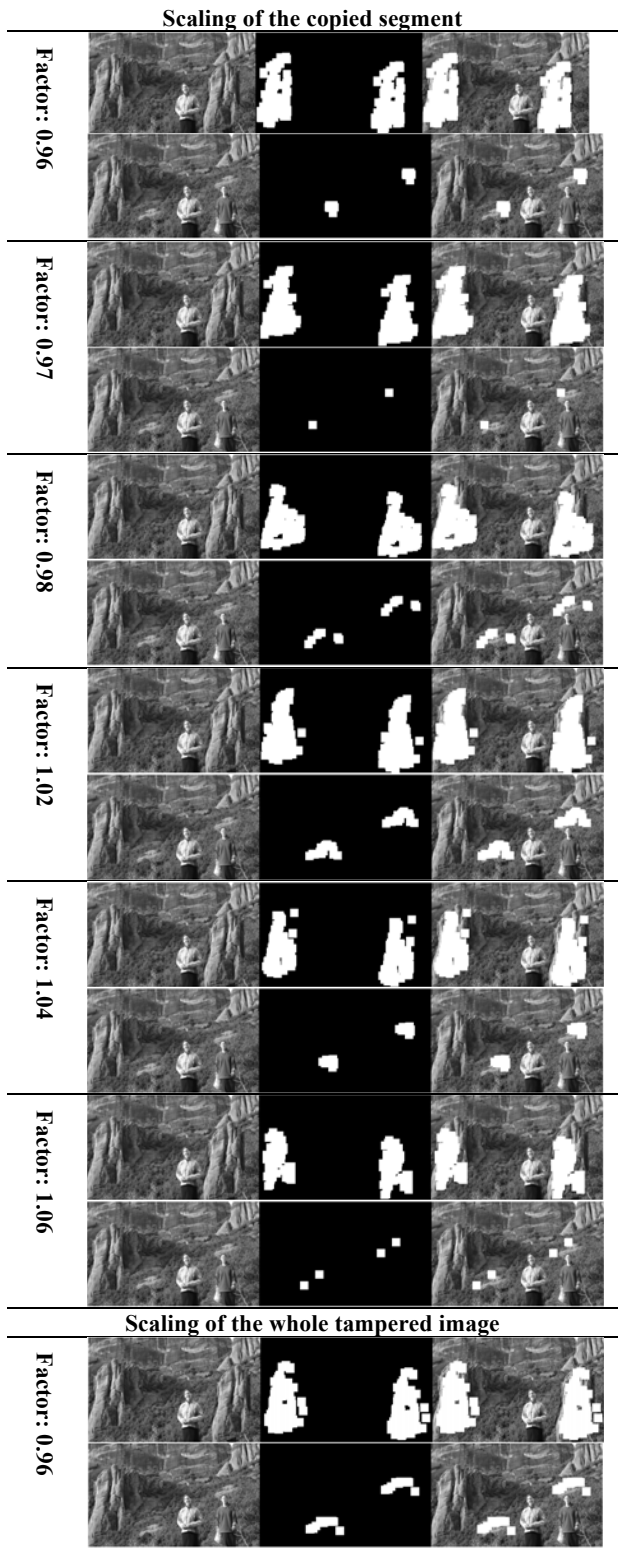

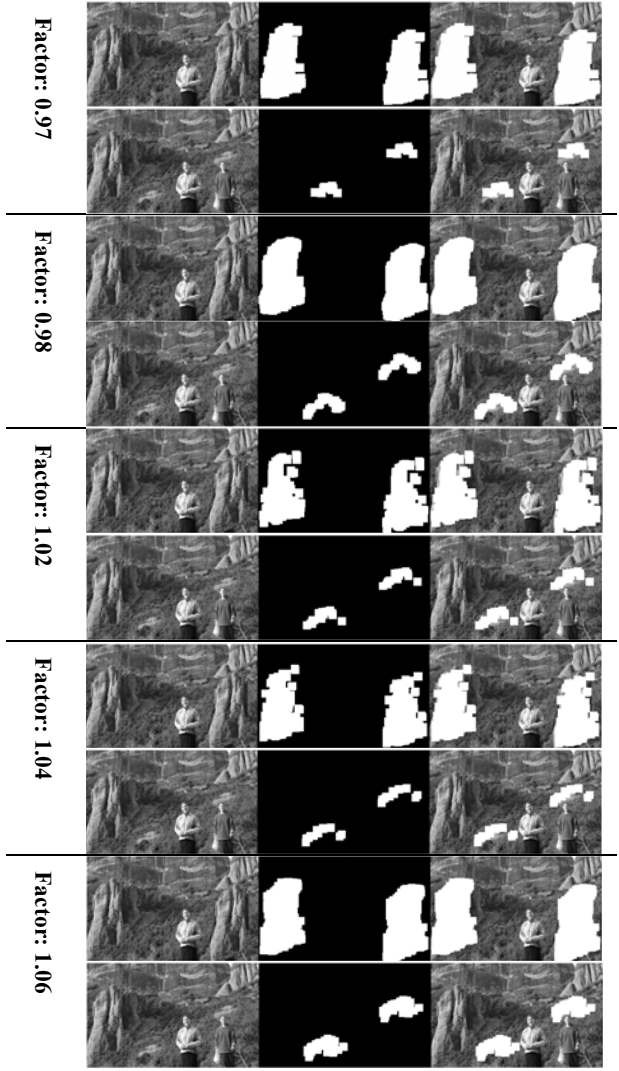

Figure 7: Results for Scaling

## 3.5 Shift

Here, the copied region is vertically/horizontally shifted by 1, 2 or 3 pixels and then pasted. Experimental results can be seen in Tab.5 and Fig.8

Table 5: Results for Shift

| Amount of pixels | Forgery 2 | | Forgery 3 | |
|---|---|---|---|---|
| | d | f | d | f |
| *Horizontal Shift (HS)* | | | | |
| 1 | 76.0608 | 4.7314 | 26.3229 | 0.7175 |
| 2 | 80.3690 | 8.5350 | 11.4350 | 0.0448 |
| 3 | 43.0819 | 2.0781 | 0 | 0 |
| *Vertical Shift (VS)* | | | | |
| 1 | 96.6359 | 15.7623 | 95.9193 | 33.5874 |
| 2 | 96.5925 | 15.4368 | 96.7713 | 36.4574 |
| 3 | 96.0608 | 15.4639 | 96.0090 | 30 |

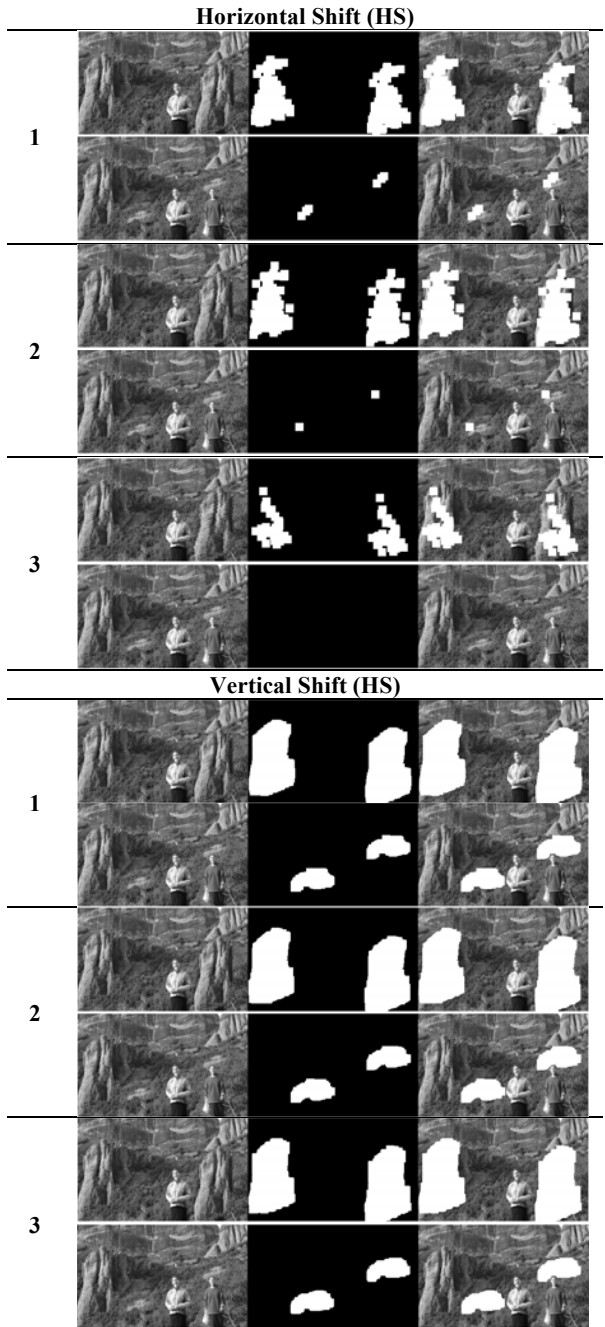

Figure 8: Results for Scaling

### 3.6 Rotation

We rotated the copied region of 2 degrees and pasted it. Tab.6 and Fig.9 show the result of the experiment.

Table 6: Results for Rotation

| Forgery 1 | | Forgery 2 | | Forgery 3 | |
|---|---|---|---|---|---|
| d | f | d | f | d | f |
| 42.7962 | 5.2003 | 61.3131 | 3.8036 | 53.8565 | 1.7937 |

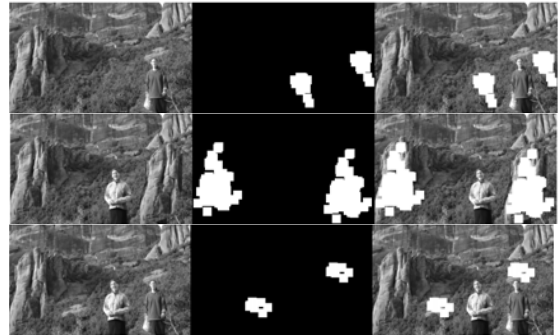

Figure 9: Results for Rotation

### 3.6 Multiple copy-move forgeries

Fig. 10 presents the outcome in case there is more than one pair of duplicated regions, precisely two pairs (Fig. 3, Forgery 4).

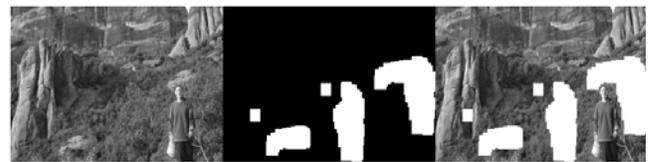

Figure 10: Results for Rotation

## 4. Conclusion

In this paper, we have presented an algorithm that can detect copy move forgeries. Our algorithm falls under the category of passive methods because it does not require any prior information on the suspicious image to proceed. Moreover, experimental results show it can detect even multiple copy move forgeries in the same image and also is relatively robust to some common distortions.

### Acknowledgments

This work is supported by the NSFC (61232016, 61173141, 61173142, 61173136, 61103215, 61070196, 61070195, and 61073191), National Basic Research Program 973 (2011CB311808), 2011GK2009, GYHY201206033, 201301030, 2013DFG12860 and PAPD fund.

## References

[1] S. Katzenbeisser, F. Petitcolas, "Information Hiding: Techniques for Steganography and Digital Watermarking", Artech House, 2000.

[2] S. Bayram, H. T. Sencar, N. Memon, "A Survey of Copy-Move Forgery Detection Techniques", In proceedings of the IEEE Western New York Image Processing Workshop, pp. 538-542, 2008.

[3] J. Fridrich, D. Soukalm, J. Lukáš, "Detection of copy-move forgery in digital images", In proceedings of the Digital Forensic Research Workshop, Cleveland, pp. 19–23, 2003.
[4] Alin C. Popescu, H. Farid, "Exposing Digital Forgeries By Detecting Duplicated Image Regions", Technical Report TR2004-515, Dartmouth College, 2004.
[5] Li Kang, Xiao-pin Cheng, "Copy-move forgery detection in digital image", 3rd International Congress on Image and Signal Processing (CISP), vol. 5, pp. 2419 – 2421, 2010.
[6] Weiqi Luo, Jiwu Huang, Guoping Qiu, "Robust Detection of Region-Duplication Forgery in Digital Images", In proceedings of the International Conference on Pattern Recognition, Washington, DC, pp. 746-749, 2006.
[7] Weihai Li and Nenghai Yu, "Rotation Robust Detection of Copy-move Forgery", In proceedings of the IEEE 17th International Conference on Image Processing, Hong Kong, 2010.
[8] Yanping Huang, Wei Lu, Wei Sun and Dongyang Long, "Improved DCT-based Detection of Copy-Move Forgery in Images", Forensic Science International, vol. 206, pp. 178-184, 2011.
[9] Yanjun Cao, Tiegang Gao, Li Fan, Qunting Yang, "A robust detection algorithm for copy-move forgery in digital images", Forensic Science International, vol. 214, pp.33–43, 2012.
[10] http://www.cs.albany.edu/~xypan/research/duplication/main.html

**Nathalie Diane Wandji** received the BSc degree in Applied Mathematics from the University of Dschang, Cameroon, in 2003; the BSc degree in Computer Science from the University of Dschang, Cameroon, in 2004; and the MSc degree in Computer Science from the University of Yaoundé 1, Cameroon, in 2008. She is currently a PhD candidate student at Hunan University, China. Her research interests include Digital Image and Video Forensics, Digital Watermarking, Information hiding and Network Security.

**Sun Xingming** received the BSc degree in Mathematics from Hunan Normal University, China, in 1984; the MSc degree in Computing Science from Dalian University of Science and Technology, China, in 1988; and the PhD degree in Computing Science from Fudan University, China, in 2001. Professor in School of Computer and Communication, Hunan University, China from 2003 to 2011, he is currently a Professor in the School of Computer and Software, Nanjing University of Information Science & Technology, China. His research interests include Network & Information Security, Digital Watermarking, Wireless Sensor Network Security, and Information hiding.

**Moise Fah Kue** is an Independent scholar. He received the BSc degree in Applied Mathematics from the University of Dschang, Cameroon, in 2001; the BSc degree in Computer Science from the University of Dschang, Cameroon, in 2002; and the DIPES II(Secondary and High School Teacher's Diploma grade II) from the ENS, University of Yaoundé 1, Cameroon, in 2006. He is currently a Pedagogical Inspector in charge of computer science for Secondary Schools in the East Region, Cameroon.